\documentclass{styles/svproc}

\usepackage{url}
\usepackage{graphicx}
\usepackage{multicol}
\usepackage{footmisc}
\usepackage{booktabs}
\usepackage{amsmath}
\usepackage{amssymb}
\usepackage{csquotes}
\usepackage{url}
\usepackage{hyperref}
\usepackage{multirow}
\usepackage{color, colortbl}
\usepackage{algorithm}
\usepackage{algpseudocode} 
\usepackage[nolist]{acronym} 

\usepackage[inline]{enumitem} 
\usepackage{cite}
\usepackage{siunitx}

\newacro{dnn}[DNN]{Deep Neural Network}
\newacro{fcn}[FCN]{Fully Convolutional Network}
\newacro{sdf}[SDF]{Signed Distance Function}
\newacro{cnn}[CNN]{Convolutional Neural Network}
\newacro{gnn}[GNN]{Graph Neural Network}
\newacro{dl}[DL]{Deep Learning}
\newacro{ml}[ML]{Machine Learning}
\newacro{gpis}[GPIS]{Gaussian Process Implicit Surface}
\newacro{mlp}[MLP]{Multi-Layer Perceptron}
\newacro{dof}[DoF]{Degree of Freedom}

\newacro{ai}[AI]{Artificial Intelligence}
\newacro{llm}[LLM]{Large Language Model}
\newacro{hri}[HRI]{Human-Robot Interaction}
\newacro{shri}[sHRI]{Social Human-Robot Interaction}

\newacro{slam}[SLAM]{Simultaneous Localization and Mapping}
\newacro{pf}[PF]{Particle Filter}
\newacro{kf}[KF]{Kalman Filter}

\newacro{ros}[ROS]{Robot Operating System}
\newacro{gui}[GUI]{Graphical User Interface}
\newacro{urdf}[URDF]{Unified Robot Description Format}

\newacro{ik}[IK]{Inverse Kinematics}

\newacro{rrmc}[RRMC]{Resolved-Rate Motion Control}
\newacro{yarp}[YARP]{Yet Another Robot Platform}

\newcommand{\equationref}[1]{\hyperref[#1]{Eq.~\ref*{#1}}}
\newcommand{\figref}[1]{\hyperref[#1]{Fig.~\ref*{#1}}}
\newcommand{\tabref}[1]{\hyperref[#1]{Table~\ref*{#1}}}
\newcommand{\secref}[1]{\hyperref[#1]{Section~\ref*{#1}}}
\newcommand{\algoref}[1]{\hyperref[#1]{Alg.~\ref*{#1}}}

\newcommand{\etal}[1]{#1 \textit{et al.}}

\newif\ifuseVspace

 \usepackage[firstpageonly=true]{draftwatermark}

\SetWatermarkAngle{0}
\SetWatermarkColor{black}
\SetWatermarkLightness{0.5}
\SetWatermarkFontSize{9pt}
\SetWatermarkVerCenter{30pt}
\SetWatermarkText{\parbox{30cm}{%
\centering This is the authors' final version of the manuscript published as:\\
\centering Rustler, L. \& Hoffmann, M. (2026),\\
\centering Learning with pyCub: A Simulation and Exercise Framework for Humanoid Robotics \\
\centering  in '17th International Conference on Robotics in Education (RiE 2026)'. (C) Springer \\
\centering
}}

\begin{document}
\mainmatter              

\title{Learning with pyCub: A Simulation and Exercise Framework for Humanoid Robotics}

\titlerunning{Learning with pyCub}
\author{Lukas Rustler \and Matej Hoffmann}

\authorrunning{Rustler, Hoffmann} 

\institute{Department of Cybernetics, Faculty of Electrical Engineering, Czech Technical University in Prague\\
\email{lukas.rustler@fel.cvut.cz, matej.hoffmann@fel.cvut.cz}}

\maketitle       

\begin{abstract}
We present pyCub, an open-source physics-based simulation of the humanoid robot iCub, along with exercises to teach students the basics of humanoid robotics. Compared to existing iCub simulators (iCub SIM, iCub Gazebo), which require C++ code and YARP as middleware, pyCub works without YARP and with Python code. The complete robot with all articulations has been simulated, with two cameras in the eyes and the unique sensitive skin of the iCub comprising 4000 receptors on its body surface. The exercises range from basic control of the robot in velocity, joint, and Cartesian space to more complex tasks like gazing, grasping, or reactive control. The whole framework is written and controlled with Python, thus allowing to be used even by people with small or almost no programming practice. The exercises can be scaled to different difficulty levels. We tested the framework in two runs of a course on humanoid robotics. The simulation, exercises, documentation, Docker images, and example videos are publicly available at \url{https://rustlluk.github.io/pyCub}.
\end{abstract}

\section{Introduction}
There are several good robotics textbooks, such as \cite{spong2006robot} or \cite{lynch2017modern} with a great video supplement, but few of them offer hands-on exercises, toolboxes, or robot simulators. The book of Corell et al.~\cite{correll2022introduction} is accompanied by a GitHub repository with Matlab and Mathematica code.  
Peter Corke developed a Python Robotics Toolbox~\cite{corke2021NotYourGrandmothers}. This toolbox provides tools for both dynamics and kinematics of robots and is a successor to the Robotic Toolbox written in MATLAB~\cite{corke1996RoboticsToolboxMATLAB}. Tutorials on robot differential kinematics using the Python Robotics Toolbox were presented by Haviland and Corke~\cite{haviland2024ManipulatorDifferentialKinematics, haviland2024ManipulatorDifferentialKinematics2}. 
\etal{Sakai}~\cite{sakai2018PythonRoboticsPythonCode} proposed an educational textbook with Python examples to explain several robotic algorithms for autonomous navigation. One part of our work is similar in the sense that we provide exercises that should teach core robotics.

\begin{figure}[htb]
    \centering
    \includegraphics[width=0.49\textwidth]{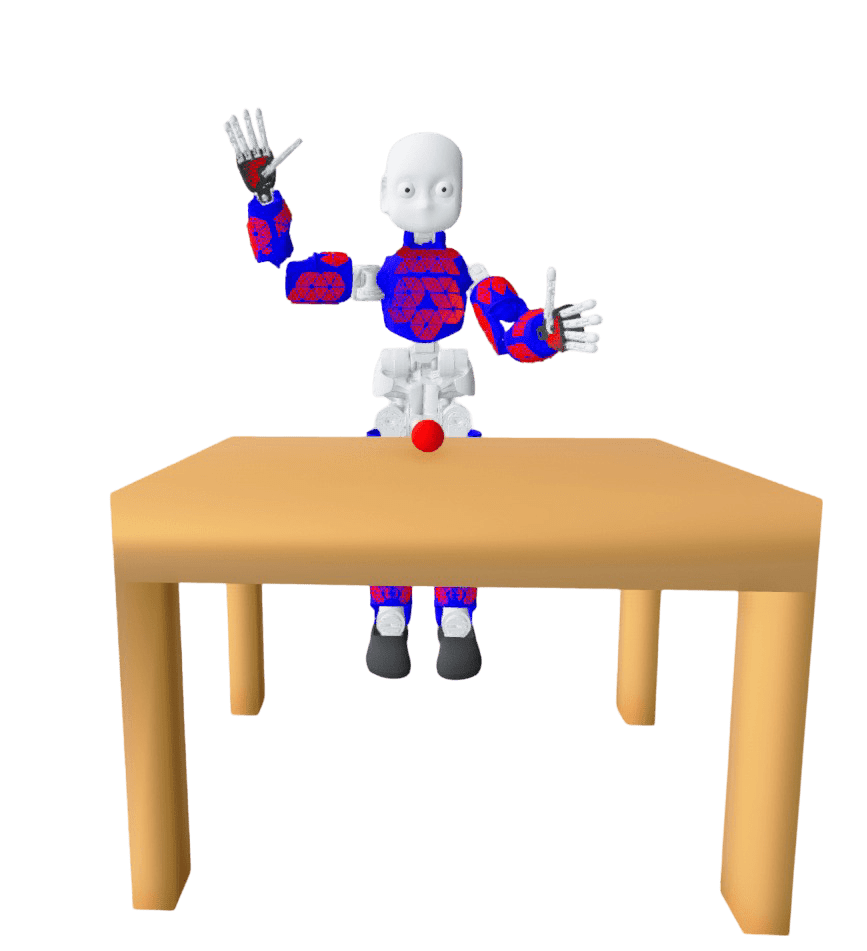}
    \includegraphics[width=0.49\textwidth]{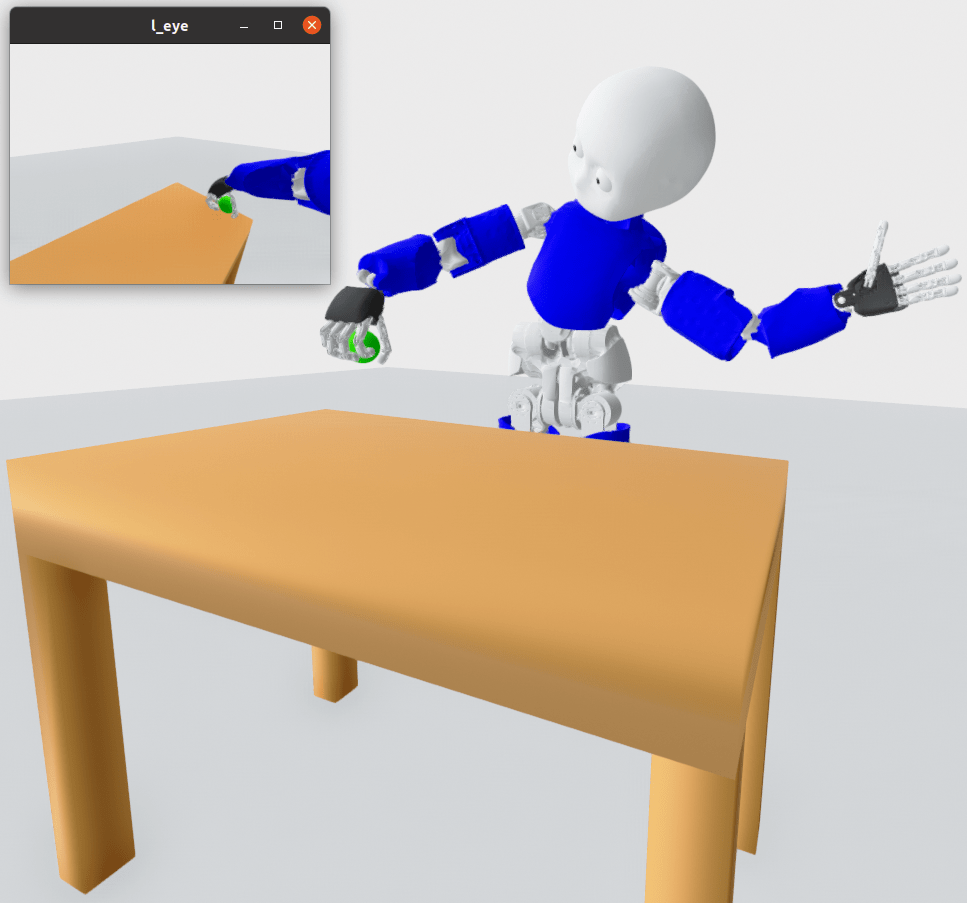}
    
    \caption{An example of the simulation environment showing the iCub humanoid robot, a table, a green ball that can be manipulated by the robot, and a robotic skin (red points in the left image). The output from the left eye (in the right image) shows a successful grasp.}
    \label{fig:grasp}
    \vspace{-2em}
\end{figure}

There are other toolboxes for kinematics and dynamics~\cite{lee2018DARTDynamicAnimation, nadeau2019PyboticsPythonToolbox} or self-calibration of robots~\cite{nadeau2019PyboticsPythonToolbox, rozlivek2020Calibration}. As many robots are controlled and unified under \ac{ros}, there are frameworks based on this middleware. The authors in~\cite{mengacci2021OpenSourceROSGazeboToolbox} created a \ac{ros} toolbox for simulating robots with compliant actuators. \etal{Fischer}~\cite{fischer2022RoboStackTutorialUsing} proposed a framework that combines \textit{Conda} and \textit{Jupyter Notebooks} for a more accessible introduction to \ac{ros}. Another interesting project is \textit{Unibotics}~\cite{roldan-alvarez2024UniboticsOpenROSbased}, which provides a web-based learning platform mainly for navigation and mapping. Furthermore, there are more specialized toolboxes, e.g., for modular industrial robots~\cite{kulz2023TimorPythonToolbox} or aerial robotic manipulators~\cite{varela-aldas2024OpenAccessPlatformSimulation}. Recently, a framework containing simulation of multiple manipulators, mobile and humanoid robots was proposed to unify robot learning~\cite{geng2025roboverseunifiedplatformdataset}.

Our presented framework was created as a tool for our university course. Our course is on humanoid robots, but due to the universal nature of such a platform, it can be used to teach several robotic topics, from kinematics and dynamics, to walking, gaze control, grasping, and human-robot interaction. Our objective thus was to create a robot simulator that is easily accessible to students and can be used for hands-on exercises in the course. We selected iCub~\cite{metta2010ICubHumanoidRobot} as it is a full humanoid and an open-source research platform. The dimensions of the robot resemble a four-year-old child, and the geometry and kinematics are designed to mimic the human body as much as possible (at the joint level, not at the muscle level). There are 53 active Degrees of Freedom (DoF). A unique feature of iCub is the whole-body artificial skin with more than 4000 individual tactile sensors. The robot has a legacy simulator ``iCub SIM''~\cite{tikhanoff2008OpensourceSimulatorCognitive}, a newer simulator in \textit{Gazebo}~\cite{mingohoffman2014YarpBasedPlugins} and offers its own training course~\cite{ICubTrainingGithub}. However, getting the environment running is an intricate process, and the default control is in \textit{C++} using the \ac{yarp}~\cite{metta2006YARPAnotherRobot} middleware, which is not suitable for beginners---there are optional Python bindings available, but as these are created from the \textit{C++} code, the syntax can be hard to grasp for less experienced users---others tackled this by creating their own wrappers, e.g., \cite{pyicubsim}. After running two editions of the course using the iCub Gazebo, even with Docker containers and other infrastructure prepared, we have evaluated the feedback forms and realized that the students complained about the complexity of the whole system. Thus, we decided to simplify the whole pipeline and create a simulator using Python, which is a more user-friendly language. The comparison of existing iCub simulators is in \tabref{tab:sim_comp}---the original simulators and a neuromorphic (and event-based) simulator~\cite{NeuromorphicIcub}. We are not aiming to replace the existing simulators, but to provide a more accessible way to iCub and humanoid robotics in general.

\begin{table}[htb]
\vspace{-1.5em}
\centering
\caption{Comparison of pyCub with existing iCub simulators (iCub SIM~\cite{tikhanoff2008OpensourceSimulatorCognitive}, iCub Gazebo~\cite{mingohoffman2014YarpBasedPlugins}, and neuromorphic iCub~\cite{NeuromorphicIcub}).}
\label{tab:sim_comp}
\resizebox{\textwidth}{!}{%
\begin{tabular}{@{} l l c c c c l l @{}}
\toprule
& \textbf{\begin{tabular}[c]{@{}l@{}}Programming\\ Language\end{tabular}} & \textbf{Middleware} & \textbf{\begin{tabular}[c]{@{}c@{}}Transferable\\ to real iCub\end{tabular}} & \textbf{Skin} & \textbf{\begin{tabular}[c]{@{}c@{}}High-level\\ planners\end{tabular}} & \textbf{\begin{tabular}[c]{@{}l@{}}Physics \\ Engine\end{tabular}} & \textbf{\begin{tabular}[c]{@{}l@{}}Rendering\\ Engine\end{tabular}} \\ 
\midrule
\textbf{iCub SIM} & \begin{tabular}[c]{@{}l@{}}C++ \\ with wrappers\end{tabular} & YARP & Yes & Yes & Yes\textsuperscript{b} & ODE & OpenGL \\ \addlinespace
\textbf{iCub Gazebo} & \begin{tabular}[c]{@{}l@{}}C++ \\ with wrappers\end{tabular} & YARP & Yes & Yes\textsuperscript{a} & Yes\textsuperscript{b} & DART\textsuperscript{c} & OGRE \\ \addlinespace
\textbf{Neuromorphic iCub} & Python & None & No & Yes & No & MuJoCo & OpenGL \\ \addlinespace
\textbf{pyCub} & Python & None & No & Yes & No & Bullet & \begin{tabular}[c]{@{}l@{}}Open3D\\ (OpenGL)\end{tabular} \\ 
\bottomrule
\multicolumn{8}{@{}l}{\footnotesize \textsuperscript{a} Skin to Gazebo was added later by the community.} \\
\multicolumn{8}{@{}l}{\footnotesize \textsuperscript{b} The planners are coming from the connection with YARP.} \\
\multicolumn{8}{@{}l}{\footnotesize \textsuperscript{c} By default, can be changed.}
\end{tabular}%
}
\vspace{-2em}
\end{table}

We present \textit{pyCub}, a Python physics-based simulator of the humanoid robot iCub. The simulator contains all limbs of the robot, together with the possibility to visualize the view from the robot's cameras, and with simulated artificial whole-body skin consisting of more than 4000 individual tactile sensors. The framework provides easy-to-use functions to control the robot in joint (position and velocity) and Cartesian space, and effortless access to proprioceptive, visual, and tactile sensors of the robot. Everything is written and accessible in Python, allowing even less advanced users to operate. The environment is fully customizable and allows to insert other objects---see \figref{fig:grasp} for an example. In addition to the simulator, we provide exercises of different difficulty levels ranging from basic control and movements to tasks typical for humanoid robots such as gazing, grasping, or whole-body skin processing and obstacle avoidance. The exercises can be further customized and allow difficulty adjustments and thus are suitable for learning (humanoid) robotics on different difficulty levels. The framework was deployed in two runs of a course on humanoid robotics.

The code, documentation, videos, and exercises are available at \url{https://rustlluk.github.io/pyCub}.

\section{Simulation Environment}
The first component of our work is the physics-based simulation environment. This Section details the implementation, customization options, and the most important methods for controlling the robot. 

\subsection{Technical Details.}

\begin{figure}[htb]
    \centering
    \includegraphics[width=0.6\textwidth]{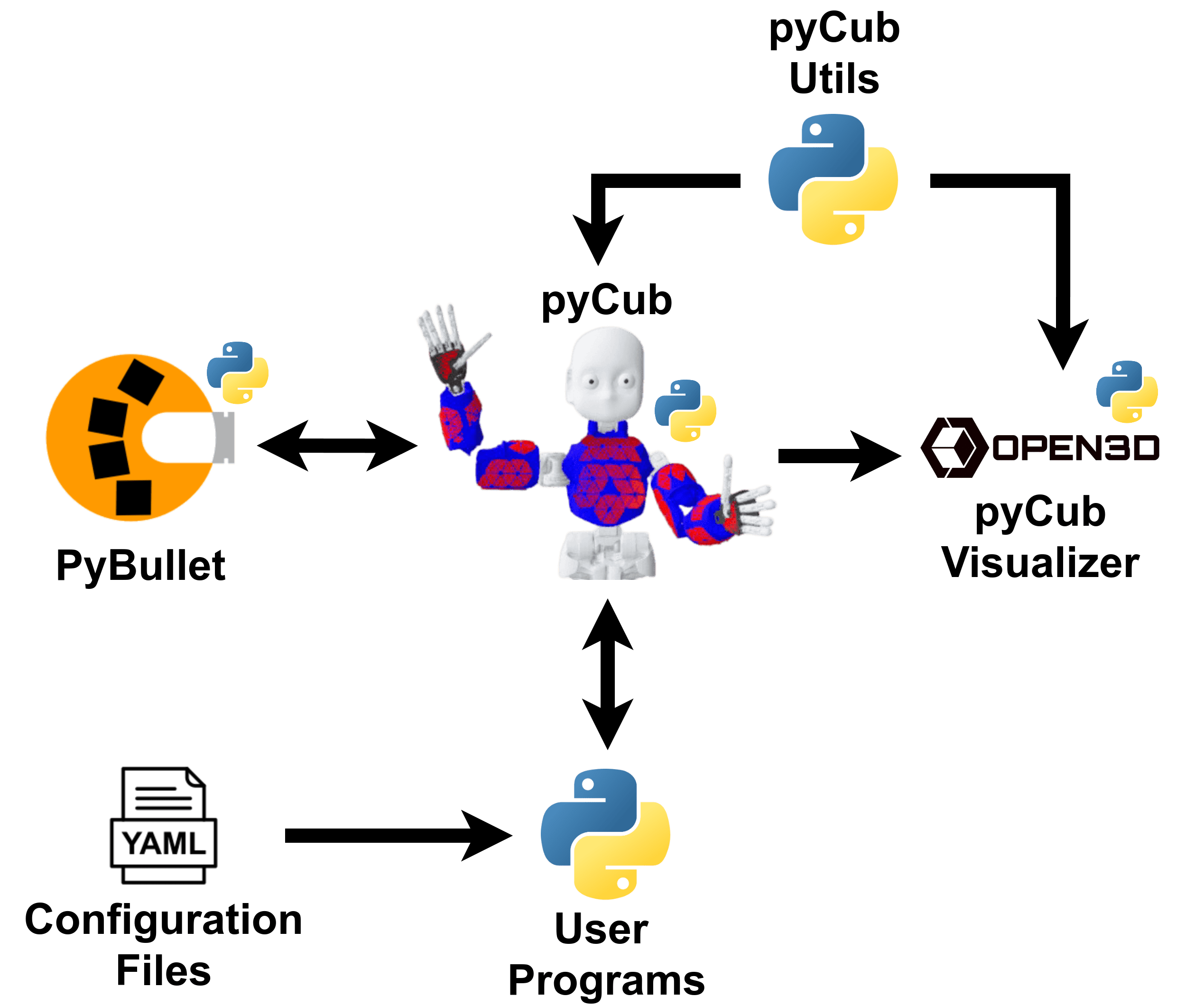}

    \caption{Schematic of the simulation package and interconnections between individual modules. The arrowheads indicate the direction of communication.}
    \label{fig:schema}
    \vspace{-2em}
\end{figure}

The physics-backbone of the simulator is realized with the \textit{PyBullet} engine~\cite{coumans2021}. This module was selected because it is open-source, multi-platform and does not require excessive computational resources---mainly compared to recent physical simulators such as Orbit (NVIDIA Issac Sim)~\cite{mittal2023orbit}. The schematic of the simulator package is shown in \figref{fig:schema}. The full documentation (and API) can be found at \url{https://rustlluk.github.io/pyCub/documentation/}.

We tested the performance (the ratio of simulation time and real time; the bigger the better) on an average laptop and a workstation, and the results are in \tabref{tab:timing}. We tested two models: 
\begin{enumerate*}[label=(\roman*)]
    \item a full model with detailed meshed and full fingers;
    \item a simplified model with less detailed meshes and without fingers.
\end{enumerate*}
We can see that on a laptop for the full model with skin and \ac{gui} enabled, the simulation runs around real-time (0.95 in this case, but the performance is based on the number of skin events). With the simple model, it increases to around 1.5. This performance is enough for the use case of the framework as one wants to see the robot moving in real-time or even slower. Without skin, performance is always higher than real time. For the workstation, the performance is always higher---\textit{PyBullet} uses a single core for all computations, and the efficiency of single-core computations makes the difference between the laptop and the workstation. In addition, tasks such as reinforcement learning could be achieved by running more instances on more cores.

\begin{table}[htb]
\centering
\caption{Average real-time factors (the ratio of simulation time and real time; the bigger the better) for two models (full model (FM) and a simple model (SM)) for four configurations---with and without the \ac{gui} with and without the skin enabled. The results are averaged over 10 repetitions. Tested on an average laptop CPU (LAP) Ryzen 7 PRO 4750u and a workstation CPU (WS) Intel i9-11900.}
\label{tab:timing}
\begin{tabular}{@{} l c c c c @{}}
\toprule
& \textbf{GUI} & \textbf{Headless} & \textbf{GUI with skin} & \textbf{Headless with skin} \\ 
\midrule
\textbf{FM - LAP} & $1.37 \pm 0.02$ & $2.03 \pm 0.02$ & $0.95 \pm 0.01$ & $1.07 \pm 0.02$ \\ 
\textbf{SM - LAP} & $6.73 \pm 0.07$ & $8.95 \pm 0.11$ & $1.46 \pm 0.05$ & $1.90 \pm 0.08$ \\ \addlinespace
\textbf{FM - WS}  & $2.25 \pm 0.00$ & $2.41 \pm 0.00$ & $1.29 \pm 0.01$ & $1.41 \pm 0.00$ \\ 
\textbf{SM - WS}  & $10.82 \pm 0.27$ & $11.46 \pm 0.03$ & $2.58 \pm 0.02$ & $2.77 \pm 0.02$ \\ 
\bottomrule
\end{tabular}
\vspace{-1em}
\end{table}

\subsubsection{Time Management.}
\vspace{-1em}
The simulation is not running by itself, but the users have to perform every step manually, i.e., call \textit{update\_simulation()}. It is advantageous because even when using less powerful computers, there is enough time to check the state of the world and perform computations between simulation steps, i.e., the visualization may run slower but everything will be computed correctly. Additionally, there is an option to actually perform the simulation step only if a predefined duration elapsed from the last one. It is useful to slow down the time to, for example, check movements in more detail. The \ac{gui} has its own fixed time management, as it is required to catch mouse events---zoom, rotations, etc.

\subsubsection{\ac{gui}.}
\vspace{-1em}
The visualization part is achieved with the \textit{Open3D} library~\cite{Zhou2018}. \textit{PyBullet} has its own built-in \ac{gui}, but it is not customizable, slow, and does not support custom simulation stepping. Therefore, we developed a new \ac{gui} that removes all the drawbacks. In addition, it allows one to easily visualize things that are not part of the simulation, e.g., visualization of gaze vectors---see \figref{fig:gui}. In native mode, \ac{gui} runs as a standalone window. However, for compatibility reasons, a web-based version can be turned on. Both versions have the possibility to view output from the eyes and allow to save RGB and depth images to a file.

\subsubsection{Customizations.}
\vspace{-1em}
The environment is fully customizable through \textit{.yaml} configuration files. The files are loaded when initializing an instance of the simulator, and thus users can create different configurations for different exercises/tasks/demos. The basic options allow to enable/disable \ac{gui}, artificial skin, eye-cameras output, logging, or self collisions. Further, the configurations allow us to set an initial position for every joint of the robot, the end-effector used for Cartesian control, or the simulation step size. Finally, users can insert additional objects into the simulation. The configurations support \ac{urdf} files, or directly \textit{.obj} triangular-mesh files. One can further specify the initial position, color, and whether the object is subject to gravity and can be moved (e.g, ball) or not (e.g., table).

\subsubsection{Installation.}
\vspace{-1em}
To provide easy installation, the framework can be installed from PyPI. Alternatively, we provide several versions of prebuilt Docker images or a version runnable entirely online using the GitPod\footnote{\url{https://www.gitpod.io/}} platform. Everything is available at \url{https://rustlluk.github.io/pyCub}. The platform was tested on several Linux distributions, Windows 10 and 11, and several versions of macOS---the ease of installation differ between systems and given computers.

\begin{figure}[htb]
    \centering
    \includegraphics[width=0.5\textwidth]{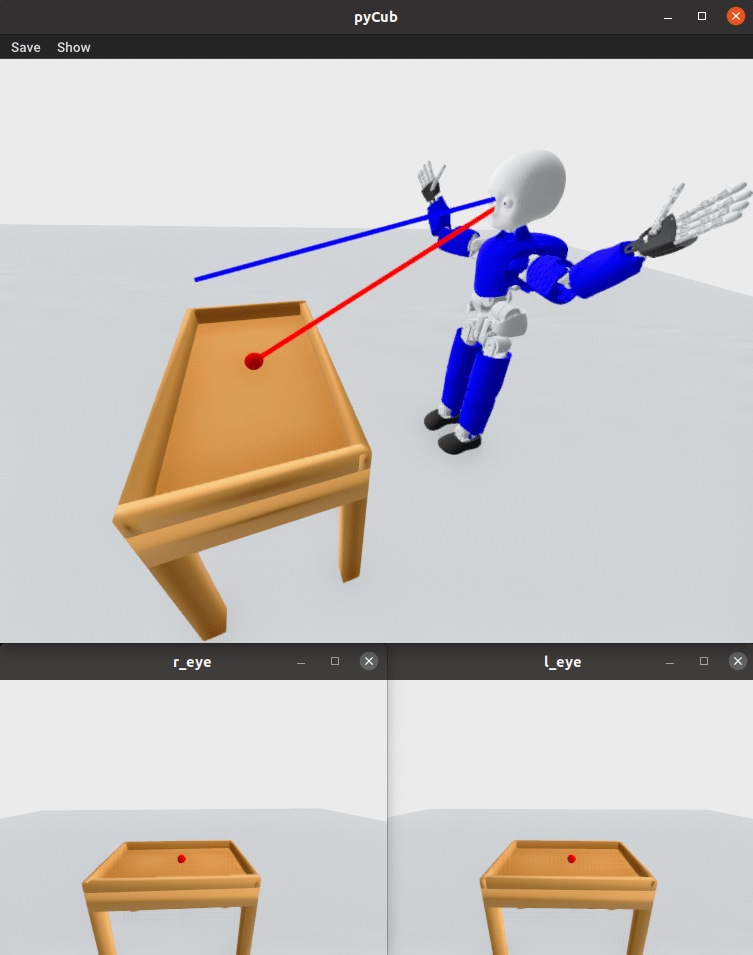}

    \caption{Example of native \ac{gui} with output from both eyes and additional visualization showing vectors where the robot is looking (blue) and where it should be looking (red) used in the gaze exercise.}
    \label{fig:gui}
    \vspace{0em}
\end{figure}

\subsection{Control Modes}
\label{sec:control_modes}
The robot can be controlled in three different modes:
\begin{enumerate*}[label=(\roman*)]
    \item joint velocity;
    \item joint position;
    \item Cartesian position.
\end{enumerate*}

\textbf{Joint Velocity Control.} The most direct and low-level is the velocity control. It directly sets the joint velocity (in \SI{}{\radian\per\second}) of every controlled joint. It is useful for reactive control. In this mode, the robot moves until a different command is set by the user. 

\textbf{Joint Position Control.} The position control sets the target joint position of each controlled joint (in \SI{}{\radian}). In this mode, the robot moves until all joints reach their set-point or, optionally, until a collision is detected. The users can additionally select whether to care about self-collision---collisions of parts of the robots.

\textbf{Cartesian Position Control.} The Cartesian mode allows to move the selected end-effector to a given 6D pose (3D position and 3D orientation). Internally, the function computes \ac{ik} for the given pose and calls position control with joint positions computed using \ac{ik}---the \ac{ik} in \textit{PyBullet} uses Damped Least Squares method with null-space support. There is no underlying planner, i.e., there are no checks whether the pose is reachable or whether the path is collision free. However, it creates a great platform for learning how movements without a planner or path smoothing look. 

\subsection{Artificial Skin}

\begin{figure}[htb]
    \centering
    \includegraphics[width=0.6\textwidth]{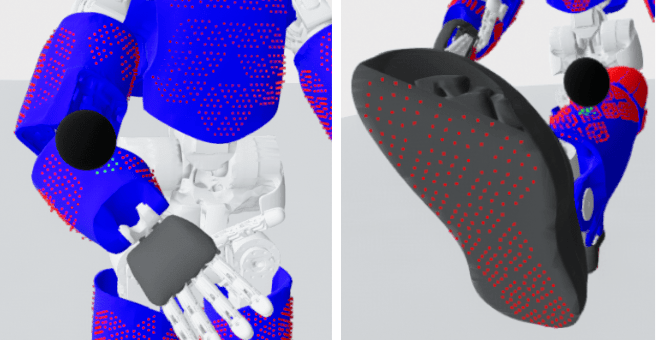}
    \caption{Examples of skin activations (green points) when a ball is touching the skin. The view is rotated such that the area \enquote{under} the ball is visible.}
    \label{fig:skin_act}
    \vspace{-3em}
\end{figure}

The artificial skin is an integral component of the robot. However, the real robot contains more than 4000 individual sensors, making the simulation challenging. The sensors are also not binary, i.e., they can return a value with 8-bit resolution. We decided to simulate the skin using raycasting. From every skin sensor (taxel) we cast a short ray with a given maximum length. If the ray hits something (other part of the robot or object in the environment), the output of the sensor is computed proportionally to the length of the ray. Using this method, we are able to simulate the intensity of contact and also the fact that, for the real skin, even taxels around the touch point are lightly activated because of the deformation of the skin. The raycasting needs to be done in every step of the simulation, and, even though the raycasting is optimized and done in batch, it is still computationally demanding. Thus, we decided to first check (for every skin part) whether there is a close object (using overlaps of bounding boxes of objects) and perform the raycasting only if necessary. The values for each activated taxels are saved in a dictionary inside the simulator and binary activations are shown as green points in the visualization---see \figref{fig:skin_act}.

\section{Exercises}
This Section introduces and describes the exercises available in the framework. We prepared several exercises that, in our opinion, span important tasks in robotics. We will describe those in recommended order based on their difficulty. However, the exercises can be further adjusted to change the difficulty level. The exercises always contain a configuration file, an assignment script, and a testing script. Videos of each exercise are available at 
\url{https://rustlluk.github.io/pyCub}.

\subsection{Push the Ball!}
The first exercise is purely for the purpose of learning how to control the simulator. The goal is to get a ball as far from a table as possible---see \figref{fig:grasp} for visualization. Basically, every imaginable approach is allowed. Whether hitting, pushing, or grasping and throwing the ball. Upon completion, users should be able to control the robot in some of the modes and understand basic functionalities. 

The testing script for this exercise computes the distance between a ball and a table after two seconds after the last movement of the robot and provides the distance in meters.

\subsection{Smooth Movements}
As described in \secref{sec:control_modes}, the Cartesian interface of the simulator does not include any underlying planner. This exercise aims to provide insight into how to move in a straight line or how to \enquote{draw} a shape, e.g., a circle. In addition, the movements should be \enquote{smooth}, i.e., a continuous movement without any jerks.

The principle is primarily in understanding that achieving a linear movement is more complicated than just setting the final 6D pose. Further, the users need to prepare the desired movement in advance (e.g., parametric equation of a circle) and sample it such that the end-effector follows it properly. Finally, the users should learn that to achieve a non-jerky movement, the sampled path points should be preempted, i.e., a new movement must be initialized before the previous one ends to avoid decelerating and accelerating between waypoints.

The testing script computes the distance from the trajectory performed to a reference one. There are different metrics for line and circle movements. However, the smoothness of the movements must be checked manually for now.

\subsection{Gaze}
Gaze control is a key ability so that the robot can bring some objects into the field of view. The goal of gaze can be described as focusing on the object of interest using, mainly, joints in the head and eyes. There is a gaze control implemented in the real iCub robot~\cite{roncone2016Cartesian6DoFGaze} that is capable of fixing to a given 3D point using all joints in the neck and eyes. In this exercise, the problem is simplified to fixating the view of the robot using neck joints to a ball moving in XY-plane. 

In the default version of the exercise, the users are given two vectors: 
\begin{enumerate*}[label=(\roman*)]
    \item vector where the robot is looking right now;
    \item vector from the robot to a ball, i.e., where the robot should be looking.
\end{enumerate*}
See \figref{fig:gui} for visualization. The task is then to use these two vectors to fixate on the ball. 

The two vectors should lead users to use fundamental algebra to find an angle $\theta$ between two vectors, for example, as:

\begin{equation}
    \theta = \arctan2\left( \left\| \mathbf{x} \times \mathbf{y} \right\|, \mathbf{x} \cdot \mathbf{y} \right),
\end{equation}

where $\mathbf{x},\mathbf{y}$ are the two vectors. The absolute value of $\theta$ is then also used as the error metric in the exercise. To complete the exercise, it is required to further decide which joint to move, e.g., by identifying the plane in which the error occurs and assigning a joint to a given plane. It can be done, for example, as the cross product $\mathbf{x} \times \mathbf{y}$, which results in the normal of the plane. Finally, to obtain the direction of movement, one can use the sign of a dot product between the plane normal and the axis of movement of the given joint.

The described solution is only one of many and can be solved in both simpler and more complicated ways. In addition, the thresholds for the allowed error can be adjusted, which can make some simpler solutions infeasible.

The testing script computes the error as the absolute value of the angle between the two vectors. The users are then given a plot of the errors in time and a score based on the mean and maximum error.

\subsection{Reactive Control and Touch Processing}
The fourth exercise focuses on reactive control in the case of contact with the environment. It is an important aspect of \ac{hri}, since ideal humanoid robots should be able to operate in close contact with humans, which inevitably leads to collisions. 

Specifically, the task is to move the robot's limbs away from contact using the information from the artificial skin. This exercise contains signal processing in the form of obtaining information from the artificial skin and finding the location of the touch based on the number of activated receptors (taxels). The users are in every step of the simulation given the positions and normals of every activated taxel for every skin part. They should cluster the points and use the biggest cluster as the touch location---one location for every activated skin part. Based on the points in the touch location and their normals, they should utilize \ac{rrmc}~\cite{whitneyRRMC} to move away, i.e., to move with the normal of the touch. \ac{rrmc} is an algorithm used for reactive behavior using velocity control. Given 6D Cartesian velocity $\dot{\mathbf{x}}$ of the end-effector (here generated using the normal of the contact), and manipulator Jacobian $\mathbf{J(q)}$ one can compute the desired joint velocities $\dot{\mathbf{q}}$ as
\begin{equation}
    \dot{\mathbf{q}} = \mathbf{J(q)}^{*}\dot{\mathbf{x}},
\end{equation}
where $\mathbf{J(q)}^{*}$ can be either a transposition $\mathbf{J(q)}^{T}$, an inverse $\mathbf{J(q)}^{-1}$, or a Moore-Penrose pseudoinverse $\mathbf{J(q)}^{+}$. One of the educational points of the exercise is to test the options, as each of them has its pros and cons. The classic inverse can only be used when the Jacobian is square and has full rank. Pseudoinverse can be unstable near singular configurations and move the end-effector in a straight line. Transpose is easier to compute, but does not necessarily move in a straight line. In the exercise, some limbs are in a configuration in which not all options provide the desired movement.

The exercise has different levels of difficulty depending on the limb in contact and the number of simultaneous contacts. The testing script here runs the different levels of difficulty. At the moment, there is no automatic evaluation, and performance must be checked manually.

\subsection{Grasping}
Robotic grasping has changed a lot during its development. From grasping based on friction cone and wrenches, through physics-based grasping to \acf{ml} and \acf{dl}-based approaches~\cite{desouzaRoboticGraspingWrench2021}. 

This exercise shows \enquote{core} grasping and is simplified by the fact that we always want to grasp the same green ball---the only thing that changes is the position of the ball on a table. The aim is also to practice core computer vision, and thus the users are guided to find the ball in the image plane (in an RGB image from one of the eyes) and obtain 2D center of the ball. The center can be deprojected to 3D given the center (u, v pixels in an image), the depth value for the tuple of pixels, and camera parameters (width; height; focal length). Then, the robot should move above the ball, get fingers to correct start position, move toward the ball, grasp it, and move it up to prove a stable grasp. An example of successful grasp can be seen in \figref{fig:grasp}.

Again, the difficulty of the exercise can be adjusted. For example, based on the information provided to the students, on the criterion for a stable grasp, or on the ball position on the table.

The testing script computes the distance of the ball from the table in meters after the hand is lifted.

\section{Conclusion, Discussion, and Future Work}
We presented \textit{pyCub}, a unified educational platform consisting of a physics-based simulation of the humanoid robot iCub and exercises that aim to teach core robotic tasks and skills. The whole framework is open-source and is available at \url{https://rustlluk.github.io/pyCub}.
The exercises cover basic robot control in joint and Cartesian space, more complicated Cartesian movements, gazing at a moving target, reactive movements in velocity space to avoid obstacles detected using signal processing from whole-body artificial skin, and grasping based on computer vision techniques.

The simulation and exercises were deployed in two runs of a course on humanoid robotics. Previous runs were taught using iCub's original Gazebo simulator controlled through a middleware \textit{YARP}~\cite{metta2006YARPAnotherRobot} using \textit{C++}. Switching to \textit{pyCub}, a Python-based simulator changed the dynamics of the class. Although the students were engineering students with experience with \textit{C++}, the whole system was troublesome to learn and most of the complaints in the post-semester survey were about \ac{yarp} and \textit{C++}. Using \textit{pyCub} allowed them to focus more on the robotic aspects of the exercises rather than on new middleware or compilation issues. For example, although the first exercise is mainly intended to get used to the simulator, the students showed that it is also an engaging exercise for them. The goal is to get the ball as far as possible from a table, and the expected solution is to simply hit the ball with an arm. However, there were many solutions such as grasping and throwing or even kicking the ball---see the accompanying video at \url{https://rustlluk.github.io/pyCub}.

A future direction is in providing a better Cartesian planner that would allow to add more exercises. An option is, for example, to create a \ac{ros} interface and then to use its Cartesian planning capabilities. We would also like to create exercises on more advanced topics such as humanoid walking or event-based sensing.

\section*{Acknowledgments}
This work was co-funded by the European Union under the project Robotics and Advanced Industrial Production (reg. no. CZ.02.01.01/00/22\_008/0004590). L.R. was additionally supported by the Grant Agency of the Czech Technical University in Prague, grant No. SGS24/096/OHK3/2T/13.

\bibliographystyle{styles/bibtex/splncs03_unsrt}
\bibliography{refs}

\end{document}